\documentclass[conference]{IEEEtran}
\IEEEoverridecommandlockouts
\usepackage{cite}
\usepackage{amsmath,amssymb,amsfonts}
\usepackage{algorithmic}
\usepackage{graphicx}
\usepackage{textcomp}
\usepackage{xcolor}
\usepackage{multirow}
\usepackage{subcaption} 
\usepackage{algorithm}
\usepackage{booktabs}
\usepackage{url}
\def\BibTeX{{\rm B\kern-.05em{\sc i\kern-.025em b}\kern-.08em
    T\kern-.1667em\lower.7ex\hbox{E}\kern-.125emX}}
\begin{document}

\title{Hyperbolic Prototype Routing for Rehearsal-Free Class-Incremental Learning}

\author{
    HongWei Zhao\textsuperscript{1},
    Rui Liu\textsuperscript{1},
    Yong Chen\textsuperscript{\dag 2}\thanks{\textsuperscript{\dag} Corresponding author.} \\[2pt]
    \textit{\textsuperscript{1}Beihang University, Beijing, China} \\ 
    \textit{\textsuperscript{2}Beijing University of Posts and Telecommunications, Beijing, China} \\[2pt]
    \{zhaohongwei, lr\}@buaa.edu.cn, alphawolf.chen@gmail.com
}

\maketitle

\begin{abstract}
Class-Incremental Learning (CIL) aims to continually learn new classes while preserving prior knowledge. Parameter-efficient fine-tuning (PEFT) with pre-trained models (PTMs) has recently shown promise, enabling CIL with minimal parameter updates. However, existing approaches still suffer from catastrophic forgetting due to cumulative interference across tasks and suboptimal module-sample matching during inference. We propose \underline{Hy}perbolic \underline{P}rototype \underline{Ro}uting (HyPro), a rehearsal-free framework that exploits hyperbolic geometry for continual learning. First, we allocate a dedicated \textbf{LoRA-Expert} module for each incremental task, enabling isolated representation learning and eliminating cross-task interference. Second, to enhance module-sample matching, we introduce a HyPro mechanism that projects features onto a Poincaré ball and performs geodesic nearest-prototype matching, yielding exponentially larger decision margins and more reliable task-level discrimination. Extensive experiments on standard CIL and Few-Shot CIL (FSCIL) benchmarks show that HyPro consistently improves average and final-stage accuracy over strong baselines. Code is available at: \url{https://github.com/Geeks-Z/ICME-HyPro-main}.

\end{abstract}


\section{Introduction}
In open-world environments, data often arrives as a continuous stream of novel categories, a paradigm formalized as \textbf{Class-Incremental Learning (CIL)}. Traditional machine learning models trained on such sequential data suffer from \textbf{catastrophic forgetting}~\cite{mccloskey1989catastrophic}, where acquiring new knowledge severely disrupts previously learned representations, leading to significant performance degradation. Incremental learning addresses this challenge by balancing the acquisition of new information with the retention of existing knowledge—a trade-off termed the \textbf{stability-plasticity dilemma}~\cite{grossberg2012studies}.

Pre-trained models (PTMs), with their strong generalization capabilities~\cite{han2021pre} derived from large-scale datasets, offer a promising foundation for CIL. However, fine-tuning all parameters of a PTM in continual learning scenarios risks undermining its inherent generalization. Recent advances mitigate this by freezing the PTM backbone and integrating parameter-efficient fine-tuning (PEFT) modules~\cite{xin2024parameter}, such as \textit{prompts}~\cite{wang2022learning,smith2023coda}, \textit{adapters}~\cite{zhou2023revisitingclassincrementallearningpretrained}, and \textit{LoRA}~\cite{liang2024inflora,wu2025sdlora}. These methods enable task-specific adaptation with minimal trainable parameters, preserving generalization while reducing forgetting.

Despite progress, two critical challenges persist:

\textbf{1. Stability--plasticity imbalance under cumulative interference}:
\begin{itemize}
    \item Shared prompt pools~\cite{wang2022learning} are prone to overwriting earlier tasks' knowledge when exposed to shifting data distributions.
    \item LoRA-based strategies~\cite{liang2024inflora,wu2025sdlora}, though effective in constraining updates to mitigate forgetting, inadvertently restrict the plasticity needed for new task adaptation.
    \item Fusion-based methods~\cite{liang2024inflora} attempt to balance old and new knowledge but often degrade the fidelity of both due to forced trade-offs in shared parameters.
\end{itemize}

\textbf{2. Inference-stage module-sample mismatches}:
\begin{itemize}
    \item Fixed PTM selection mechanisms~\cite{wang2022learning,wang2022s} struggle under substantial domain shifts, resulting in suboptimal activations and degraded predictions.
    \item With volume growth, high-dimensional features concentrate into narrow cones (``Cone Effect''~\cite{gong2018geometry}), causing prototype crowding and ambiguous boundaries as tasks accumulate, which leads to inference-stage module—sample mismatches.
    
\end{itemize}

These issues motivate our core research question: \textbf{Can we jointly optimize stability-plasticity and module retrieval to mitigate catastrophic forgetting?}

To this end, we propose \textbf{HyPro}, a novel rehearsal-free framework for PEFT-based CIL. HyPro comprises two synergistic components operating in distinct stages. First, we dynamically allocate dedicated LoRA-Experts for each new incremental task. Unlike shared or sequentially fine-tuned modules, each LoRA-Expert exclusively encodes task-specific knowledge, with only the current expert being trainable and all prior experts frozen. These lightweight experts are strategically integrated into Transformer architectures—specifically within multilayer perceptrons (MLPs) and multi-head attention (MHA) layers—to efficiently capture discriminative task-specific features. Second, to overcome the geometric limitations of Euclidean feature space, we project features into a hyperbolic manifold and perform geodesic prototype routing for module-sample. By leveraging the exponentially expanding capacity of hyperbolic space, our approach ensures large inter-class distances and improves sample-to-module alignment.

Our principal contributions are threefold:
\begin{itemize}
    \item We introduce \textbf{dynamic LoRA-Expert} allocation into PEFT-based CIL, allowing each incremental task to obtain its own dedicated expert without predefining the total number of experts. This design ensures strict parameter isolation across tasks and enhances the model’s plasticity to new knowledge.
    
    \item We propose a \textbf{hyperbolic prototype routing} mechanism that embeds features into a non-Euclidean manifold for module selection. By leveraging the exponentially expanding capacity of hyperbolic space to preserve large pairwise margins between class prototypes, this design overcomes the limitations of conventional Euclidean feature representations during inference.

    \item We provide rigorous empirical validation across six challenging benchmarks, establishing new state-of-the-art results in diverse incremental learning scenarios. Furthermore, we demonstrate architectural flexibility through HyPro-MLP, a variant employing alternative LoRA-Expert placement that maintains competitive performance while offering implementation alternatives.
\end{itemize}

\section{Related Work}
\subsection{Class Incremental Learning}
\textbf{Class-Incremental Learning} requires models to continually learn new classes while retaining prior knowledge. Existing methods can be broadly categorized as follows~\cite{wang2022learning}:

\textbf{Parameter regularization-based methods}~\cite{aljundi2019task} mitigate catastrophic forgetting by constraining important weights. However, they often underperform on complex datasets or in more challenging incremental scenarios~\cite{rebuffi2017icarl}.

\textbf{Rehearsal-based methods} utilize data replay to reinforce old knowledge, either by storing raw images~\cite{rebuffi2017icarl} or intermediate features~\cite{yu2020semantic}. While effective, these methods are limited by buffer sizes and potential privacy concerns.

\textbf{Dynamic network-based methods}~\cite{wang2022foster} allocate new sub-networks for each incremental task while freezing previous ones, thus effectively balancing stability and plasticity. However, they introduce substantial memory overhead due to continually expanding parameters, and many still require old samples for cross-network fusion.

\subsection{PEFT-Based CIL}
Recent advances in PEFT have gained prominence for enabling efficient training and inference with minimal parameters while preserving the strong generalization capabilities of PTMs. Integrating PEFT into PTM-based CIL offers a promising approach to enhancing continual learning performance.

\textbf{Prompt-based methods} like L2P~\cite{wang2022learning} introduced a shared prompt pool for sequential learning.  S-Prompts~\cite{wang2022s} learned domain-specific prompts retrieved by K-NN, while CODA-Prompt~\cite{smith2023coda} proposed an end-to-end decomposed attention framework without rehearsal.

\textbf{LoRA-based methods}, such as InfLoRA~\cite{liang2024inflora}, project updates into a gradient-orthogonal subspace to avoid interference, requiring extensive data storage. MoE-Adapters~\cite{MoE-Adapters} employed a dynamic activate-freeze strategy for inter-task collaboration, but its pre-defined number of LoRAs limited adaptation to subsequent incremental tasks. SD-LoRA~\cite{wu2025sdlora} decoupled gradient direction and magnitude, preserving early-task directions while adapting to new tasks via shrinking, at the cost of reduced plasticity.

\subsection{Geometric Deep Learning and Hyperbolic Embeddings}
\textbf{Geometric deep learning} studies representation learning beyond Euclidean vector spaces, leveraging non-Euclidean structures such as graphs and Riemannian manifolds. Among them, \textbf{hyperbolic embeddings} are particularly effective for data with latent hierarchical or tree-like organization, because hyperbolic space provides exponentially expanding volume with radius, enabling compact yet highly separable representations.

\textbf{Poincaré embeddings}~\cite{nickel2017poincare} demonstrated that hyperbolic space can represent hierarchical relations with substantially lower distortion than Euclidean embeddings. Subsequently, \textbf{hyperbolic neural networks} generalized common operations to the manifold using tools such as exponential/log maps and Möbius addition, enabling end-to-end learning in hyperbolic space~\cite{ganea2018hyperbolic,khrulkov2020hyperbolic}. In vision, hyperbolic representations have also been shown to be beneficial for capturing fine-grained relations and improving discrimination when features exhibit implicit hierarchy~\cite{khrulkov2020hyperbolic}. These properties motivate us to exploit hyperbolic geometry to enlarge inter-class margins for more reliable prototype-based routing in continual learning.

\section{Preliminaries}

\textbf{Problem Definition:} We study class-incremental learning over a stream of tasks $\{\mathcal{D}_t\}_{t=1}^{T}$, where $\mathcal{D}_t=\{(\mathbf{x}_i,\mathbf{y}_i)\}_{i=1}^{n_t}$ and task label spaces are disjoint ($\mathcal{Y}_t\cap\mathcal{Y}_{t'}=\emptyset$ for $t\neq t'$). Under the rehearsal-free setting~\cite{wang2022learning,smith2023coda}, training on task $t$ uses only $\mathcal{D}_t$ to learn $f_{\Theta}(\mathbf{x})=\mathbf{W}_{cls}^{\top}\phi(\mathbf{x})$ by minimizing
\begin{equation}
\text{L}(\mathcal{D}_t)=\frac{1}{|\mathcal{D}_t|}\sum_{(\mathbf{x}_i,\mathbf{y}_i)\in\mathcal{D}_t}\text{L}\!\left(f_{\Theta}(\mathbf{x}_i),\mathbf{y}_i\right),
\end{equation}
and evaluation after task $t$ is performed on all observed classes $\mathbf{Y}_t=\bigcup_{k=1}^{t}\mathcal{Y}_k$. The embedding function $\phi\left(\cdot\right)$ refers to the final \text{[CLS]} token in \textbf{Vision Transformer (ViT)} architecture~\cite{dosovitskiy2020image} and $\mathbf{W}_{cls}$ represents the classifier parameters.

\begin{figure*}
    \centering
    \includegraphics[width=0.9\linewidth,height=0.38\textheight]{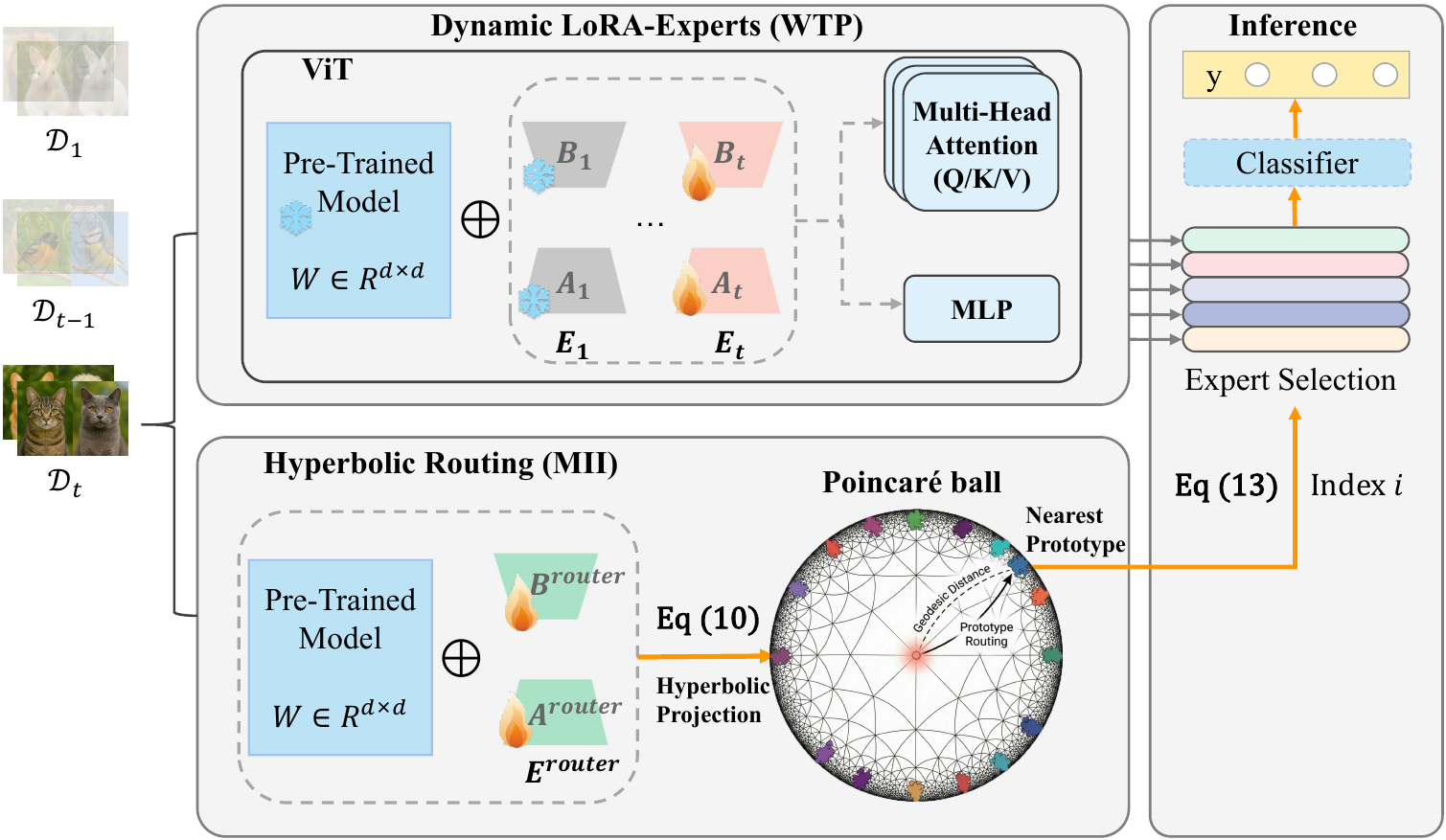}
    \caption{ Illustration of HyPro. In the $t$-th incremental task, a new LoRA-Expert $\mathbf{E}_t$ (with parameters $\mathbf{A}_t$ and $\mathbf{B}_t$) is trained to capture task-specific features. Domain-specific features from the router $\mathbf{E}^{router}$ are projected into the hyperbolic space (Poincaré ball). During inference, the nearest prototype guides LoRA-Experts selection for each input sample.
    }
    \label{HyPro}
\end{figure*}

\textbf{Low-Rank Adaptation.} LoRA was introduced to efficiently fine-tune large PTMs by injecting low-rank updates into weight matrices~\cite{hu2022lora}. Given a pre-trained weight matrix $\mathbf{W} \in \mathbf{R}^{d \times d}$, LoRA learns an additive low-rank decomposition:
\begin{equation}
\mathbf{W} + \Delta \mathbf{W} = \mathbf{W} + \mathbf{B}\mathbf{A} ,
\label{eq:lora}
\end{equation}
where $\mathbf{A} \in \mathbf{R}^{r \times d}$, $\mathbf{B} \in \mathbf{R}^{d \times r}$ and the rank $r \ll d$. This reduces trainable parameters while maintaining expressiveness. LoRA enables cost-effective, scalable fine-tuning, making it suitable for continual learning where efficiency and avoiding forgetting are critical.

\textbf{Hyperbolic Poincar\'e Ball Model.} 
We adopt the $d$-dimensional Poincar\'e ball model of constant sectional curvature $-c$ ($c>0$),
\begin{equation}
\mathbb{D}_c^d \;=\; \{\mathbf{x}\in\mathbb{R}^d \mid c\|\mathbf{x}\|^2 < 1\},
\end{equation}
i.e., the open Euclidean ball of radius $1/\sqrt{c}$. The Riemannian metric is conformal to the Euclidean metric $g^E$:
\begin{equation}
g_{\mathbf{x}}^{\mathbb{D}} \;=\; \lambda_{\mathbf{x}}^2\, g^E, \qquad
\lambda_{\mathbf{x}} \;=\; \frac{2}{1-c\|\mathbf{x}\|^2}.
\end{equation}
The geodesic distance between $\mathbf{x},\mathbf{y}\in\mathbb{D}_c^d$ admits the closed form
\begin{equation}
d_{\mathbb{D}}(\mathbf{x},\mathbf{y})
\;=\; \frac{1}{\sqrt{c}}\operatorname{arcosh}\!\Bigg(1+2c\,\frac{\|\mathbf{x}-\mathbf{y}\|^2}
{(1-c\|\mathbf{x}\|^2)(1-c\|\mathbf{y}\|^2)}\Bigg).
\label{eq:poincare_geodesic_distance_rewritten}
\end{equation}
Note that the argument of $\operatorname{arcosh}$ is $\ge 1$ and that as $\|\mathbf{x}\|\to 1/\sqrt{c}$ the conformal factor $\lambda_{\mathbf{x}}\to\infty$, so distances to the boundary diverge. Consequently, hyperbolic space exhibits exponential volume growth near the boundary; this geometric trait makes the Poincar\'e ball suitable for embedding hierarchical or tree-like structures, since a rapidly growing representational capacity is available compared to Euclidean space of the same nominal dimensionality.

\section{The Proposed Method}

Following HiDe-Prompt~\cite{wang2023hierarchical}, we decompose CIL prediction into two probabilistic components: Module-Identity Inference (MII) and Within-Task Prediction (WTP). By Bayes' theorem, the probability of a sample $\mathbf{x}$ belonging to class $j$ in task $i$ is:
\begin{align}\label{BayesTheorem}
    {P}(\mathbf{x} \in \mathcal{X}_{i,j}|\mathcal{D},\Theta) = \underbrace{{P}(\mathbf{x} \in \mathcal{X}_{i,j}|\mathbf{x} \in \mathcal{X}_{i},\mathcal{D},\Theta)}_{\text{WTP}} \underbrace{P(\mathbf{x} \in \mathcal{X}_{i}|\mathcal{D},\Theta)}_{\text{MII}}.
\end{align}
Eq.~(\ref{BayesTheorem}) implies that overall performance hinges on optimizing both WTP and MII. However, existing methods often falter in these areas: iterative updates or fusion degrade WTP via interference~\cite{wang2022learning,liang2024inflora}, while reliance on fixed PTM features limits MII under domain shifts~\cite{wang2022s}. To overcome these limitations, we propose \textbf{HyPro}, which explicitly targets both components: \textbf{Dynamic LoRA-Experts} ensure isolated, high-fidelity WTP, while \textbf{Hyperbolic Prototype Routing} leverages geometric properties to maximize MII accuracy. An overview is shown in Fig.~\ref{HyPro}.

\subsection{Dynamic LoRA-Experts}
We investigate the integration of \textbf{LoRA-Expert} into the ViT. ViT first splits an input image into fixed-size patches, which are then linearly projected and augmented with positional embeddings before being processed by the Transformer encoder. The encoder comprises multi-head self-attention (MHA) layers and multilayer perceptron (MLP) blocks. To adaptively capture \textbf{task-specific features} in incremental learning, we dynamically incorporate a \textbf{LoRA-Expert module} at each incremental stage. This module can be attached either as a parallel branch to the MLP (\textbf{MLP-Expert}) or to the query ($\mathbf{W}_q$), key ($\mathbf{W}_k$), or value ($\mathbf{W}_v$) projections in the MHA mechanism (\textbf{QKV-Expert}). When applied to both $\mathbf{W}_q$ and $\mathbf{W}_v$, the configuration is termed (\textbf{QV-Expert}). Formally, letting $\phi(\mathbf{x};\mathbf{E}_t)$ denote the embedding with LoRA-Expert $\mathbf{E}_t$, we write the MLP and MHA updates as:
\begin{equation}
    \phi(\mathbf{x};\mathbf{E}_t^{\text{MLP}})= \mathbf{e} + \text{MLP}(\mathbf{e}) + \mathbf{E}^{\text{MLP}}_{t}(\mathbf{e}),
    \label{eq:mlp_expert}
\end{equation}
\begin{equation}
    \phi(\mathbf{x};\mathbf{E}_t^{\text{QKV}})=
    \text{Attn}\!\big(
        \mathbf{u}_Q + \mathbf{E}^{Q}_{t}(\mathbf{e}),\,
        \mathbf{u}_K + \mathbf{E}^{K}_{t}(\mathbf{e}),\,
        \mathbf{u}_V + \mathbf{E}^{V}_{t}(\mathbf{e})
    \big),
    \label{eq:qkv_expert}
\end{equation}
where $\mathbf{e}$ and \(\mathbf{u}\) are the input and output of the original module, and $\mathbf{E}_t$ denotes the output of the $t$-th LoRA-Expert. Here, the superscripts (e.g., $\mathbf{E}_t^{\text{MLP}}$, $\mathbf{E}_t^{Q}$, $\mathbf{E}_t^{K}$, $\mathbf{E}_t^{V}$) indicate different fine-tuning locations within the ViT block. The attention operator is
\begin{equation}
    \text{Attn}(\mathbf{Q}, \mathbf{K}, \mathbf{V})
    = \text{softmax}\!\left(\frac{\mathbf{Q}\mathbf{K}^\top}{\sqrt{d}}\right)\mathbf{V},
    \label{eq:attn}
\end{equation}
and the multi-head formulation is omitted for brevity.

Each LoRA-Expert $\mathbf{E}_t$ introduces task-specific parameters $\mathbf{A}_t$ and $\mathbf{B}_t$, computing $\mathbf{E}_{t}(\mathbf{e}) = \mathbf{B}_t \mathbf{A}_t\, \mathbf{e}$. While the backbone remains shared, each task maintains a distinct expert. We initialize $\mathbf{B}_t$ to zero and $\mathbf{A}_t$ via Kaiming initialization~\cite{he2015delving} for the first task; subsequent experts inherit weights from the preceding task to accelerate convergence. Crucially, during training on task $t$, all prior experts ($\mathbf{E}_{1 \dots t-1}$) are frozen. This isolation strategy ensures that new knowledge is acquired without interfering with previously learned representations, effectively mitigating catastrophic forgetting with minimal parameter overhead.

\subsection{Hyperbolic Prototype Routing}
While the frozen PTM backbone offers robust generalization, it inherently lacks the plasticity to capture task-specific discriminative features, particularly under significant domain shifts. Relying solely on PTM representations for module selection often leads to suboptimal expert retrieval. To bridge this gap, we introduce a \textbf{Hyperbolic Prototype Routing} mechanism that synergizes a learnable domain adapter with the exponential capacity of the Poincaré manifold.

\subsubsection{Manifold Projection and Learnable Router}
We employ a dedicated router module, $\mathbf{E}^{router}$, which shares the architecture of LoRA-Experts but is continuously updated across all tasks. For an input $\mathbf{x}$, the router extracts a domain-specific feature $\mathbf{z} \in \mathbb{R}^d$ with $\phi(\mathbf{x};\mathbf{E}^{router})$. To mitigate the ``Cone Effect'' in high-dimensional Euclidean space, we project $\mathbf{z}$ onto the $d$-dimensional Poincaré ball $\mathbb{D}_c^d$ with curvature $-c$.
Using the exponential map at the origin, $\exp_{0}^c: T_0 \mathbb{D}_c^d \to \mathbb{D}_c^d$, the Euclidean feature is mapped to the hyperbolic embedding $\mathbf{h}$:
\begin{equation}
    \mathbf{h} = \exp_{0}^c(\mathbf{z}) = \tanh\left(\sqrt{c}\,\|\mathbf{z}\|\right) \frac{\mathbf{z}}{\sqrt{c}\,\|\mathbf{z}\|}.
    \label{eq:exp_map}
\end{equation}
The $\tanh$ nonlinearity naturally compresses the embedding magnitude, pushing highly discriminative features toward the boundary of $\mathbb{D}_c^d$. In this boundary region, the hyperbolic volume expands exponentially, providing ample geometric capacity to separate accumulating task prototypes without interference.

The router network $\mathbf{E}^{router}$ is trained after each new task’s LoRA-Expert is learned. To construct the optimization targets, we first extract domain-specific prototypes $\mathbf{P}$ from the trained LoRA-Expert of the current task.
For a task $t$ with dataset $\mathcal{D}_t$, the prototype for class $i$ is formulated as:
\begin{equation}
	\mathbf{P}_i = \frac{1}{N_i} \sum_{j=1}^{|\mathcal{D}_t|}\mathbb{I}(y_j=i)\phi(\mathbf{x}_j;\mathbf{E}_{t}).
 \label{eq:prototype}
\end{equation}
Let $\mathbf{h}_{p} = \exp_{0}^c(\mathbf{P}_{i})$ denote the hyperbolic prototype of the ground-truth class $i$.

To ensure the router accurately assigns samples to these prototypes while retaining knowledge of previous tasks, we minimize a hybrid loss defined directly on the manifold:
\begin{equation}
\min_{\mathbf{E}^{router}_{t}}{\text{H}_{router}}=\alpha d_{\mathbb{D}}\!\big(\mathbf{h}_{q},\,\mathbf{h}_{p}\big) + (1-\alpha) d_{\mathbb{D}}\!\big(\mathbf{h}_{q},\,\mathbf{h}_{old}\big),
\label{router_loss}
\end{equation}
where $\mathbf{h}_{q}$ is the router's output for input $\mathbf{x}$, and $\alpha \in $ is a trade-off hyperparameter.
The first term (plasticity) aligns the router's output with the current task's expert features. The second term (stability) acts as hyperbolic knowledge distillation by aligning the current embedding $\mathbf{h}_{q}$ with the router output from the previous stage, 
$\mathbf{h}_{old} = \exp_{0}^c\big(\phi(\mathbf{x};\mathbf{E}_{\text{router}}^{t-1})\big)$,
thereby preventing decision boundary drift for prior tasks. The computation of $d_{\mathbb{D}}$ follows Eq.~(\ref{eq:poincare_geodesic_distance_rewritten}).

\subsubsection{Expert Selection}
During inference, the query embedding $\mathbf{h}_{q}$ is matched against the stored prototype using the Riemannian geodesic distance:
\begin{equation}
i = {\text{argmin}}(d_{\mathbb{D}}(\mathbf{h}_{q}, \mathbf{h}_{p})) .
\label{eq:expert_selection}
\end{equation}
The LoRA-Expert corresponding to the nearest prototype is then activated for final prediction.
Crucially, as prototypes approach the boundary, the denominator $(1-c\|\mathbf{h}\|^2) \to 0$, causing inter-class distances to diverge. This creates large ``safety margins'' between tasks, minimizing routing confusion.

\subsection{Optimization Objective for HyPro}

The optimization process of HyPro consists of two parts: \textbf{one} is the learning of the dynamic LoRA-Expert, and its optimization objective function can be formulated as:
\begin{equation}
   \min_{\mathbf{W}_{cls}, \mathbf{E}_i}\text{L}\left(\mathbf{W}_{cls}^\top\phi\left(\mathbf{x};\mathbf{E}_i\right),\mathbf{y}\right)
    \label{ce_loss},
\end{equation}
where $\text{L}$ is the cross-entropy loss. In the second stage, the router is optimized via Eq.~(\ref{router_loss}). 

During inference, we adopt the class prototypes extracted by the LoRA-Experts as the classifier weights, $\mathbf{W}_{cls} = \mathbf{P}$, and use a cosine classifier for final prediction:
\begin{equation} \label{eq:final_fx}
\text{f}(\mathbf{x}|\mathbf{E}_i)=(\frac {\mathbf{W}_{cls}}{\|\mathbf{W}_{cls}\|_2})^\top(\frac{\phi(\mathbf{x};\mathbf{E}_i)}{\|\phi(\mathbf{x};\mathbf{E}_i)\|_2}),
\end{equation}
where $\mathbf{E}_i$ denotes the selected LoRA-Expert by Eq.~(\ref{eq:expert_selection}) for the input $\mathbf{x}$.


\section{Experiments}
\subsection{Experimental Settings}

\textbf{Datasets:}
We evaluate HyPro on two tasks: CIL and Few-Shot Class-Incremental Learning (FSCIL)~\cite{tao2020few}. For CIL, we follow standard protocols~\cite{wu2025sdlora} and test on five benchmarks: CIFAR100~\cite{krizhevsky2009learning}, CUB200~\cite{wah2011caltech}(ImageNet-R~\cite{hendrycks2021many}, Omnibenchmark~\cite{zhang2022benchmarking}, and VTAB~\cite{zhai2019large} are provided in the \textit{Supplementary}), with classes evenly divided into $T$ incremental tasks. 

For FSCIL, we adopt the settings of PriViLege~\cite{park2024pre} and ASP~\cite{liu2024few} on CUB200~(100-base 10-way 5-shot) and CIFAR100~(60-base 5-way 5-shot), where ``100-base''   indicates that the first task contains 100 classes with sufficient training samples, ``10-way 5-shot'' means that each subsequent task introduces 10 novel classes, each with only 5 examples.  

\textbf{Evaluation metrics:} For CIL, we evaluate performance using two standard metrics: the average accuracy over all incremental tasks $\bar{\mathcal{A}} = \frac{1}{T}\sum_{i=1}^{T}\mathcal{ACC}_{i}$ and the accuracy on the last task $\mathcal{A}_L$~\cite{liang2024inflora}. Here, we denote the $\text{Top-1}$ average accuracy  after training on the $i$-th task as $\mathcal{ACC}_{i}$. For FSCIL, we report the first task accuracy $\mathcal{A}_{\text{Base}}$, last task accuracy $\mathcal{A}_{L}$, and overall average accuracy $\bar{\mathcal{A}}$.

\begin{table}[t]\scriptsize
    \centering
    \caption{
        Performance comparison of selected \textbf{CIL} methods, all built on the same pre-trained backbone ({\bf ViT-B/16-IN21K}). 
    }
    \label{tab:cil_benchmarks}
    \resizebox{1.0\linewidth}{!}{
    \begin{tabular}{@{}l *{2}{cc} c@{}}
        \toprule
        \multirow{2}{*}{Method} & 
        \multicolumn{2}{c}{CIFAR100 ($T$=10)} & 
        \multicolumn{2}{c}{CUB200 ($T$=10)} & 
        \\
        \cmidrule(lr){2-3} \cmidrule(lr){4-5} 
        & $\mathcal{A}_L$ & $\bar{\mathcal{A}}$ 
        & $\mathcal{A}_L$ & $\bar{\mathcal{A}}$ \\
        \midrule
        Full Fine-Tuning 
            & $66.26$ & $76.94$ 
            & $55.29$ & $70.30$   
            \\
        
        SimpleCIL~\cite{zhou2023revisitingclassincrementallearningpretrained} 
            & $81.27$ & $87.13$ 
            & $82.28$ & $91.85$ 
            \\
        
        L2P~\cite{wang2022learning}
            & $84.82$ & $89.78$ 
            & $71.98$ & $81.80$ 
            \\
        
        CODA-Prompt~\cite{smith2023coda} 
            & $86.69$ & $91.31$ 
            & $75.45$ & $84.65$ 
             \\
        InfLoRA~\cite{liang2024inflora} 
            &$86.43$ &$91.80$ 
            &$70.07$ &$81.71$ 
            \\
        
        SD-LoRA~\cite{wu2025sdlora} 
            &$87.62$ &$92.10$ 
            &$72.69$ &$83.17$ 
            \\
        MoE-Adapters~\cite{MoE-Adapters} 
            &$77.96$ &$85.19$ 
            &$52.62$ &$65.67$ 
            
            \\
        \midrule
        HyPro-MLP 
            &89.13&93.23
            &87.79&92.11 
            
            \\
        HyPro-QV
            &\bf{89.68}&\bf{93.62}	
            &\bf{88.13}&\bf{92.20}	
            \\
        \bottomrule
    \end{tabular}
    }
\end{table}

\textbf{Architecture and Training details:} 
We adopt ViT-B/16-IN21K~\cite{dosovitskiy2020image} as our pre-trained backbone, initialized with weights from ImageNet-21K. For optimization, we use SGD with an initial learning rate of 0.02, decayed via a cosine annealing schedule. The LoRA-Expert modules are trained for 20 epochs with a batch size of 48, while the router undergoes 5 epochs of training under the same batch size configuration. LoRA decomposition employs a rank of $r$=4, curvature $c=0.5$, and the plasticity feature distillation coefficient $\alpha$ is set to 0.1. To ensure reproducibility, all experiments are conducted on an NVIDIA A800 GPU using identical data splits and pre-trained backbones. Reported results are averaged over three independent runs to account for variability. 

For more experimental details, please see the \textit{Supplementary} and the forthcoming \textit{code} release on GitHub.

\begin{table}[t]\scriptsize
    \centering
    \caption{Performance comparison of selected \textbf{FSCIL} methods, all built on the same pre-trained backbone ({\bf ViT-B/16-IN21K}).
        }
    \label{tab:fscil_benchmarks}
    \resizebox{1.0\linewidth}{!}{
        \begin{tabular}{@{}lccccccccc@{}}
            \toprule
            \multirow{2}{*}{Method} & 
            \multicolumn{3}{c}{CUB200 ($T$=11)} & 
            \multicolumn{3}{c}{CIFAR100 ($T$=9)} &
            \\
            \cmidrule(lr){2-4} \cmidrule(l){5-7} 
            & $\mathcal{A}_{\text{Base}}$ & $\mathcal{A}_L$ & $\bar{\mathcal{A}}$ 
            & $\mathcal{A}_{\text{Base}}$ & $\mathcal{A}_L$ & $\bar{\mathcal{A}}$\\
            \midrule
        
        L2P~\cite{wang2022learning} 
            & $91.50$ & $50.04$ & $66.70$ 
            & $93.43$ & $55.75$ & $71.81$
            \\
        
        CODA-Prompt~\cite{smith2023coda}
            & $91.50$ & $53.65$ & $69.30$ 
            & $94.05$ & $57.10$ & $73.11$
            \\
        
        InfLoRA~\cite{liang2024inflora} 
            & $92.45$ & $45.18$ & $66.27$ 
            & $\bf{94.92}$ & $57.41$ & $74.28$
            \\
        
        SD-LoRA~\cite{wu2025sdlora} 
            & $91.92$ & $56.28$ & $70.87$ 
            & $94.60$ & $73.51$ & $78.42$
            \\
        \midrule
        CPE-CLIP~\cite{10350931}
            & $80.21$ & $63.32$ & $69.37$ 
            & $88.32$ & $79.99$ & $83.38$
            \\
        ASP~\cite{liu2024few}
            & $87.14$ & \bf{82.86} & $83.46$ 
            & $91.77$ & $86.04$ & $88.54$
            \\
        PriViLege~\cite{park2024pre} 
            & $82.21$ & $75.08$ & $77.50$ 
            & $90.88$ & $86.06$ & $88.08$
            \\
        \midrule
        HyPro-MLP 
            &92.68 &76.84 &\bf{84.92} 
            &93.75 &\bf{88.96} &\bf{91.16}
            \\
        HyPro-QV 
            &\bf{93.35} &66.50 &82.48 
            &93.80 &88.08 &90.54
            \\
            \bottomrule
        \end{tabular}
        }
\end{table}

\subsection{Benchmark Comparison}
\textbf{Class Incremental Learning:} We evaluate HyPro against SOTA methods across five benchmarks. As shown in Table~\ref{tab:cil_benchmarks}, HyPro consistently achieves superior performance on CIFAR100 and CUB200, while results on ImageNet-R, Omnibenchmark, and VTAB (detailed in the \textit{Supplementary}) further confirm its efficacy. Specifically, HyPro outperforms leading LoRA-based methods (InfLoRA and SD-LoRA) by margins of 1.5\%--2.0\% on CIFAR100 and 15\%--18\% on CUB200 in terms of $\mathcal{A}_L$. Notably, both HyPro-MLP and HyPro-QV variants exhibit robust performance, maintaining a clear advantage over competing approaches.

\textbf{Few-Shot Class-Incremental Learning:}
We further evaluate HyPro on few-shot class-incremental learning. As shown in Table~\ref{tab:fscil_benchmarks}, HyPro consistently delivers superior accuracy, establishing new state-of-the-art results on multiple benchmarks. In particular, it achieves the highest last accuracy ($\mathcal{A}_L$) and average accuracy ($\bar{\mathcal{A}}$) on CUB200 and CIFAR100. For instance, on CUB200, HyPro-MLP achieves an average accuracy of 84.92\%, outperforming the prior best method, ASP, by a margin of 1.46\%. On CIFAR100, HyPro-MLP reaches 91.16\%, surpassing ASP by 2.62\%.


\begin{table}[t]\scriptsize
    \centering
    \caption{Ablation studies on {\bf CIL} and {\bf FSCIL} tasks. 
        The first dataset corresponds to CIL, and the second to FSCIL. For each metric, the left/right values represent performance with MLP-LoRA and QV-LoRA fine-tuning, respectively. HPR: Hyperbolic Prototype Routing.
        }
    \label{tab: ablation_studies_components_cil}
    \resizebox{1.0\linewidth}{!}{
        \begin{tabular}{@{}l|cc|cc@{}}
            \toprule
            \multirow{2}{*}{Ablated Components} 
            & \multicolumn{2}{c|}{ImageNet-R ($T$=5)} 
            & \multicolumn{2}{c}{CIFAR100 ($T$=9)} 
            \\
            \cmidrule(lr){2-3} \cmidrule(lr){4-5} 
            & $\mathcal{A}_L$ & $\bar{\mathcal{A}}$
            & $\mathcal{A}_L$ & $\bar{\mathcal{A}}$\\
            \midrule
            w/o LoRA Dynamically 
            &61.17\ /\ 72.37 &74.31\ /\ 80.14
            &73.03\ /\ 79.73 &80.39\ /\ 86.59  
            \\
            w/o HPR 
            &69.53\ /\ 73.13 &77.75\ /\ 80.42
            &81.13\ /\ 84.09 & 86.97\ /\ 88.67 
            \\
            \midrule
            HyPro-MLP\ /\ QV 
            &\bf{77.00\ /\ 78.10} &\bf{82.44\ /\ 83.05} 
            &\bf{88.96\ /\ 88.08 } &\bf{91.16\ /\ 90.54}
            \\
            \bottomrule
        \end{tabular} 
        }
\end{table}

\subsection{Ablation Study}
\textbf{Different Components:} 
We conduct ablation studies to assess the contribution of each component in HyPro (Table~\ref{tab: ablation_studies_components_cil}). The variant \textbf{w/o LoRA Dynamically} uses a single LoRA for all tasks, resulting in a significant performance drop under large domain shifts—indicating severe forgetting and task interference. In contrast, assigning a dedicated LoRA per task better preserves stability–plasticity trade-offs across incremental steps. This underscores the value of task-specific LoRAs in PEFT-based continual learning. Removing the hyperbolic prototype routing (\textbf{w/o HPR}) and using frozen class prototypes as keys also degrades performance, confirming the router's critical role in effective module–sample matching.

\begin{table}[t]\scriptsize
    \centering
    \caption{
        Router average accuracy comparison of different module-sample matching strategies on {\bf CIL} tasks. All methods are based on the same pre-trained backbone ({\bf ViT-B/16-IN21K}).
    }
    \label{tab:diff_router_compare}
    \resizebox{1.0\linewidth}{!}{
    \begin{tabular}{@{}lccc@{}}
        \toprule
        Method & CIFAR100 ($T$=10) & CUB200 ($T$=10) & ImageNet-R ($T$=5) \\
        \midrule
        KNN 
            & 86.80 & 90.40 & 75.24 \\
       Prototype 
            & 89.60 & 91.10 & 77.04 \\
        \midrule
        HyPro-MLP 
            & 93.71 & 93.40 & 88.48 \\
        HyPro-QV
            & 94.13 & 93.32 & 88.61 \\
        \bottomrule
    \end{tabular}
    }
\end{table}

\subsection{Further Analysis}

\textbf{Impact of Matching Strategies:} To validate the effectiveness of our routing mechanism, we compare the proposed Hyperbolic Prototype Routing against two baseline strategies: (1) \textbf{K-Nearest Neighbors (KNN, K=3)}, which retrieves experts based on raw feature similarity, and (2) \textbf{Euclidean Prototype}, which employs a learnable router identical to ours but operates within Euclidean space using cosine similarity. As presented in Table~\ref{tab:diff_router_compare}, HyPro significantly outperforms all baselines. The superior performance over the Euclidean counterpart confirms that projecting features onto the Poincaré manifold effectively alleviates the ``Cone Effect'', providing larger decision margins for more accurate expert retrieval. 

\section{Conclusion}
We propose HyPro, a novel framework that mitigates catastrophic forgetting by embedding task-specific features into a Poincar\'e ball manifold. By exploiting the exponential capacity of hyperbolic space, HyPro resolves the ``Cone Effect'' prevalent in Euclidean PEFT methods. Extensive experiments confirm that HyPro establishes new state-of-the-art results across CIL and FSCIL benchmarks. Future work will explore hyperbolic operations for inter-layer feature fusion.


\begin{thebibliography}{10}
\providecommand{\url}[1]{#1}
\csname url@samestyle\endcsname
\providecommand{\newblock}{\relax}
\providecommand{\bibinfo}[2]{#2}
\providecommand{\BIBentrySTDinterwordspacing}{\spaceskip=0pt\relax}
\providecommand{\BIBentryALTinterwordstretchfactor}{4}
\providecommand{\BIBentryALTinterwordspacing}{\spaceskip=\fontdimen2\font plus
\BIBentryALTinterwordstretchfactor\fontdimen3\font minus \fontdimen4\font\relax}
\providecommand{\BIBforeignlanguage}[2]{{%
\expandafter\ifx\csname l@#1\endcsname\relax
\typeout{** WARNING: IEEEtran.bst: No hyphenation pattern has been}%
\typeout{** loaded for the language `#1'. Using the pattern for}%
\typeout{** the default language instead.}%
\else
\language=\csname l@#1\endcsname
\fi
#2}}
\providecommand{\BIBdecl}{\relax}
\BIBdecl

\bibitem{mccloskey1989catastrophic}
M.~McCloskey and N.~J. Cohen, ``Catastrophic interference in connectionist networks: The sequential learning problem.''\hskip 1em plus 0.5em minus 0.4em\relax Academic Press, 1989, vol.~24, pp. 109--165.

\bibitem{grossberg2012studies}
S.~T. Grossberg, \emph{Studies of mind and brain: Neural principles of learning, perception, development, cognition, and motor control}.\hskip 1em plus 0.5em minus 0.4em\relax Springer Science \& Business Media, 2012, vol.~70.

\bibitem{han2021pre}
X.~Han, Z.~Zhang, N.~Ding, Y.~Gu, X.~Liu, Y.~Huo, J.~Qiu, Y.~Yao, A.~Zhang, L.~Zhang \emph{et~al.}, ``Pre-trained models: Past, present and future,'' \emph{AI Open}, vol.~2, pp. 225--250, 2021.

\bibitem{xin2024parameter}
Y.~Xin, S.~Luo, H.~Zhou, J.~Du, X.~Liu, Y.~Fan, Q.~Li, and Y.~Du, ``Parameter-efficient fine-tuning for pre-trained vision models: {A} survey,'' \emph{CoRR}, vol. abs/2402.02242, 2024.

\bibitem{wang2022learning}
Z.~Wang, Z.~Zhang, C.-Y. Lee, H.~Zhang, R.~Sun, X.~Ren, G.~Su, V.~Perot, J.~Dy, and T.~Pfister, ``Learning to prompt for continual learning,'' in \emph{2022 CVPR}, 2022, pp. 139--149.

\bibitem{smith2023coda}
J.~S. Smith, L.~Karlinsky, V.~Gutta, P.~Cascante-Bonilla, D.~Kim, A.~Arbelle, R.~Panda, R.~Feris, and Z.~Kira, ``Coda-prompt: Continual decomposed attention-based prompting for rehearsal-free continual learning,'' in \emph{2023 CVPR}, 2023, pp. 11\,909--11\,919.

\bibitem{zhou2023revisitingclassincrementallearningpretrained}
D.-W. Zhou, Z.-W. Cai, H.-J. Ye, D.-C. Zhan, and Z.~Liu, ``Revisiting class-incremental learning with pre-trained models: Generalizability and adaptivity are all you need,'' \emph{IJCV}, pp. 1--21, 2024.

\bibitem{liang2024inflora}
Y.-S. Liang and W.-J. Li, ``Inflora: Interference-free low-rank adaptation for continual learning,'' in \emph{2024 CVPR}, 2024, pp. 23\,638--23\,647.

\bibitem{wu2025sdlora}
Y.~Wu, H.~Piao, L.-K. Huang, R.~Wang, W.~Li, H.~Pfister, D.~Meng, K.~Ma, and Y.~Wei, ``Sd-lora: Scalable decoupled low-rank adaptation for class incremental learning,'' in \emph{2025 ICLR}, 2025.

\bibitem{wang2022s}
Y.~Wang, Z.~Huang, and X.~Hong, ``S-prompts learning with pre-trained transformers: An occam's razor for domain incremental learning,'' in \emph{Advances in Neural Information Processing Systems 35}, 2022.

\bibitem{gong2018geometry}
\BIBentryALTinterwordspacing
J.~Gao, D.~He, X.~Tan, T.~Qin, L.~Wang, and T.~Liu, ``Representation degeneration problem in training natural language generation models,'' in \emph{International Conference on Learning Representations}, 2019. [Online]. Available: \url{https://openreview.net/forum?id=SkEYojRqtm}
\BIBentrySTDinterwordspacing

\bibitem{aljundi2019task}
R.~Aljundi, K.~Kelchtermans, and T.~Tuytelaars, ``Task-free continual learning,'' in \emph{2019 CVPR}, 2019, pp. 11\,246--11\,255.

\bibitem{rebuffi2017icarl}
S.-A. Rebuffi, A.~Kolesnikov, G.~Sperl, and C.~H. Lampert, ``icarl: Incremental classifier and representation learning,'' in \emph{2017 CVPR}, 2017, pp. 5533--5542.

\bibitem{yu2020semantic}
L.~Yu, B.~Twardowski, X.~Liu, L.~Herranz, K.~Wang, Y.~Cheng, S.~Jui, and J.~van~de Weijer, ``Semantic drift compensation for class-incremental learning,'' in \emph{2020 CVPR}, 2020, pp. 6980--6989.

\bibitem{wang2022foster}
F.~Wang, D.~Zhou, H.~Ye, and D.~Zhan, ``{FOSTER:} feature boosting and compression for class-incremental learning,'' in \emph{2022 ECCV}, ser. Lecture Notes in Computer Science, S.~Avidan, G.~J. Brostow, M.~Ciss{\'{e}}, G.~M. Farinella, and T.~Hassner, Eds., vol. 13685.\hskip 1em plus 0.5em minus 0.4em\relax Springer, 2022, pp. 398--414.

\bibitem{MoE-Adapters}
J.~Yu, Y.~Zhuge, L.~Zhang, P.~Hu, D.~Wang, H.~Lu, and Y.~He, ``Boosting continual learning of vision-language models via mixture-of-experts adapters,'' in \emph{2024 CVPR}, 2024, pp. 23\,219--23\,230.

\bibitem{nickel2017poincare}
M.~Nickel and D.~Kiela, ``Poincar\'{e} embeddings for learning hierarchical representations,'' in \emph{Proceedings of the 31st International Conference on Neural Information Processing Systems}, ser. NIPS'17.\hskip 1em plus 0.5em minus 0.4em\relax Red Hook, NY, USA: Curran Associates Inc., 2017, p. 6341–6350.

\bibitem{ganea2018hyperbolic}
O.-E. Ganea, G.~B\'{e}cigneul, and T.~Hofmann, ``Hyperbolic neural networks,'' in \emph{Proceedings of the 32nd International Conference on Neural Information Processing Systems}, ser. NIPS'18.\hskip 1em plus 0.5em minus 0.4em\relax Red Hook, NY, USA: Curran Associates Inc., 2018, p. 5350–5360.

\bibitem{khrulkov2020hyperbolic}
V.~Khrulkov, L.~Mirvakhabova, E.~Ustinova, I.~Oseledets, and V.~Lempitsky, ``Hyperbolic image embeddings,'' in \emph{2020 CVPR}, June 2020.

\bibitem{dosovitskiy2020image}
A.~Dosovitskiy, L.~Beyer, A.~Kolesnikov, D.~Weissenborn, X.~Zhai, T.~Unterthiner, M.~Dehghani, M.~Minderer, G.~Heigold, S.~Gelly, J.~Uszkoreit, and N.~Houlsby, ``An image is worth 16x16 words: Transformers for image recognition at scale,'' in \emph{2021 ICLR}, 2021.

\bibitem{hu2022lora}
E.~J. Hu, Y.~Shen, P.~Wallis, Z.~Allen{-}Zhu, Y.~Li, S.~Wang, L.~Wang, and W.~Chen, ``Lora: Low-rank adaptation of large language models,'' in \emph{2022 ICLR}, 2022.

\bibitem{wang2023hierarchical}
L.~Wang, J.~Xie, X.~Zhang, M.~Huang, H.~Su, and J.~Zhu, ``Hierarchical decomposition of prompt-based continual learning: Rethinking obscured sub-optimality,'' in \emph{Advances in Neural Information Processing Systems 36: Annual Conference on Neural Information Processing Systems 2023, NeurIPS 2023, New Orleans, LA, USA, December 10 - 16, 2023}, A.~Oh, T.~Naumann, A.~Globerson, K.~Saenko, M.~Hardt, and S.~Levine, Eds., 2023.

\bibitem{he2015delving}
K.~He, X.~Zhang, S.~Ren, and J.~Sun, ``Delving deep into rectifiers: Surpassing human-level performance on imagenet classification,'' in \emph{2015 ICCV}, 2015, pp. 1026--1034.

\bibitem{tao2020few}
X.~Tao, X.~Hong, X.~Chang, S.~Dong, X.~Wei, and Y.~Gong, ``Few-shot class-incremental learning,'' in \emph{2020 CVPR}, 2020, pp. 12\,180--12\,189.

\bibitem{krizhevsky2009learning}
A.~Krizhevsky and G.~Hinton, ``Learning multiple layers of features from tiny images,'' 2009.

\bibitem{wah2011caltech}
C.~Wah, S.~Branson, P.~Welinder, P.~Perona, and S.~Belongie, ``The caltech-ucsd birds-200-2011 dataset,'' 2011.

\bibitem{hendrycks2021many}
D.~Hendrycks, S.~Basart, N.~Mu, S.~Kadavath, F.~Wang, E.~Dorundo, R.~Desai, T.~Zhu, S.~Parajuli, M.~Guo \emph{et~al.}, ``The many faces of robustness: A critical analysis of out-of-distribution generalization,'' in \emph{Proceedings of the IEEE/CVF international conference on computer vision}, 2021, pp. 8340--8349.

\bibitem{zhang2022benchmarking}
Y.~Zhang, Z.~Yin, J.~Shao, and Z.~Liu, ``Benchmarking omni-vision representation through the lens of visual realms,'' in \emph{European Conference on Computer Vision}.\hskip 1em plus 0.5em minus 0.4em\relax Springer, 2022, pp. 594--611.

\bibitem{zhai2019large}
X.~Zhai, J.~Puigcerver, A.~Kolesnikov, P.~Ruyssen, C.~Riquelme, M.~Lucic, J.~Djolonga, A.~S. Pinto, M.~Neumann, A.~Dosovitskiy \emph{et~al.}, ``A large-scale study of representation learning with the visual task adaptation benchmark,'' \emph{arXiv preprint arXiv:1910.04867}, 2019.

\bibitem{park2024pre}
K.-H. Park, K.~Song, and G.-M. Park, ``Pre-trained vision and language transformers are few-shot incremental learners,'' in \emph{2024 CVPR}, 2024, pp. 23\,881--23\,890.

\bibitem{liu2024few}
C.~Liu, Z.~Wang, T.~Xiong, R.~Chen, Y.~Wu, J.~Guo, and H.~Huang, ``Few-shot class incremental learning with attention-aware self-adaptive prompt,'' in \emph{2024 ECCV}.\hskip 1em plus 0.5em minus 0.4em\relax Cham: Springer Nature Switzerland, 2024, pp. 1--18.

\bibitem{10350931}
M.~D’Alessandro, A.~Alonso, E.~Calabrés, and M.~Galar, ``Multimodal parameter-efficient few-shot class incremental learning,'' in \emph{2023 ICCVW}, 2023, pp. 3385--3395.

\end{thebibliography}

\vspace{12pt}
\color{red}

\end{document}